\documentclass[runningheads]{llncs}

\usepackage{eccv}

\usepackage{eccvabbrv}

\usepackage{graphicx}
\usepackage{booktabs}

\usepackage[accsupp]{axessibility}  

\usepackage{amssymb}            
\usepackage{mathtools}          
\usepackage{mathrsfs}           
\usepackage{graphicx}           
\usepackage{subcaption}         
\usepackage[space]{grffile}     
\usepackage{url}                
\usepackage{booktabs} 
\usepackage{multirow} 
\usepackage{adjustbox}
\usepackage[font=scriptsize,labelfont=bf,aboveskip=0.5em,belowskip=0pt]{caption}
\usepackage{comment}

\usepackage{hyperref}

\usepackage{orcidlink}

\newcommand\blfootnote[1]{%
  \begingroup
  \renewcommand\thefootnote{}\footnote{#1}%
  \addtocounter{footnote}{-1}%
  \endgroup
}

\begin{document}


\title{Generating Physically Realistic and Directable Human Motions from Multi-Modal Inputs}

\titlerunning{Generating motions from multi-modal inputs}

\author{Aayam Shrestha$^{\ast}$\inst{1} \and
Pan Liu$^{\ast}$\inst{2} \and
German Ros$^{\dagger}$\inst{2}  \and
Kai Yuan$^{\ddagger}$\inst{2}  \and
Alan Fern\inst{1}
}

\authorrunning{A. Shrestha et al.}

\institute{Oregon State University \and
Intel Labs
\blfootnote{$\ast$ Equal contribution\\
$\dagger$ Now at NVIDIA \\
$\ddagger$ Corresponding author. Email: kai.yuans{\texttt{@}}intel.com}
}

\maketitle


%




\begin{abstract}
This work focuses on generating realistic, physically-based human behaviors from multi-modal inputs, which may only partially specify the desired motion.
For example, the input may come from a VR controller providing arm motion and body velocity, partial key-point animation, computer vision applied to videos, or even higher-level motion goals.  
This requires a versatile low-level humanoid controller that can handle such sparse, under-specified guidance, seamlessly switch between skills, and recover from failures. Current approaches for learning humanoid controllers from demonstration data capture some of these characteristics, but none achieve them all. To this end, we introduce the Masked Humanoid Controller (MHC), a novel approach that applies multi-objective imitation learning on augmented and selectively masked motion demonstrations. The training methodology results in an MHC that exhibits the key capabilities of \textit{catch-up} to out-of-sync input commands, \textit{combining} elements from multiple motion sequences, and \textit{completing} unspecified parts of motions from sparse multimodal input. We demonstrate these key capabilities for an MHC learned over a dataset of 87 diverse skills and showcase different multi-modal use cases, including integration with planning frameworks to highlight MHC’s ability to solve new user-defined tasks without any finetuning. Project webpage: \url{https://idigitopia.github.io/projects/mhc/}

\end{abstract}

\section{Introduction}
\label{sec:intro}

Physically simulated humanoid characters have the potential to generate natural looking and realistic human motions and behaviors for a variety of applications ranging from video games, robotics, virtual/augmented reality, to digital avatars. However, directable and natural motion generation for these physically unstable, high-dimensional characters corresponds to a challenging control problem requiring precise coordination of joint-level commands \cite{wagener_mocapact_2023, luo_perpetual_2023}. One solution is to leverage motion capture (MoCap) data, which provides detailed pose information for demonstrations of real human motion that are relevant to an application. We can then attempt to train a motion generator that can be directed to imitate the individual motion demonstrations as well as combinations or transformations of these motions. This data-driven imitation-learning approach offers the potential for producing directable behavior while ensuring human-like motion quality.

\begin{figure}[tb]
  \centering
  \includegraphics[width=1.0\linewidth]{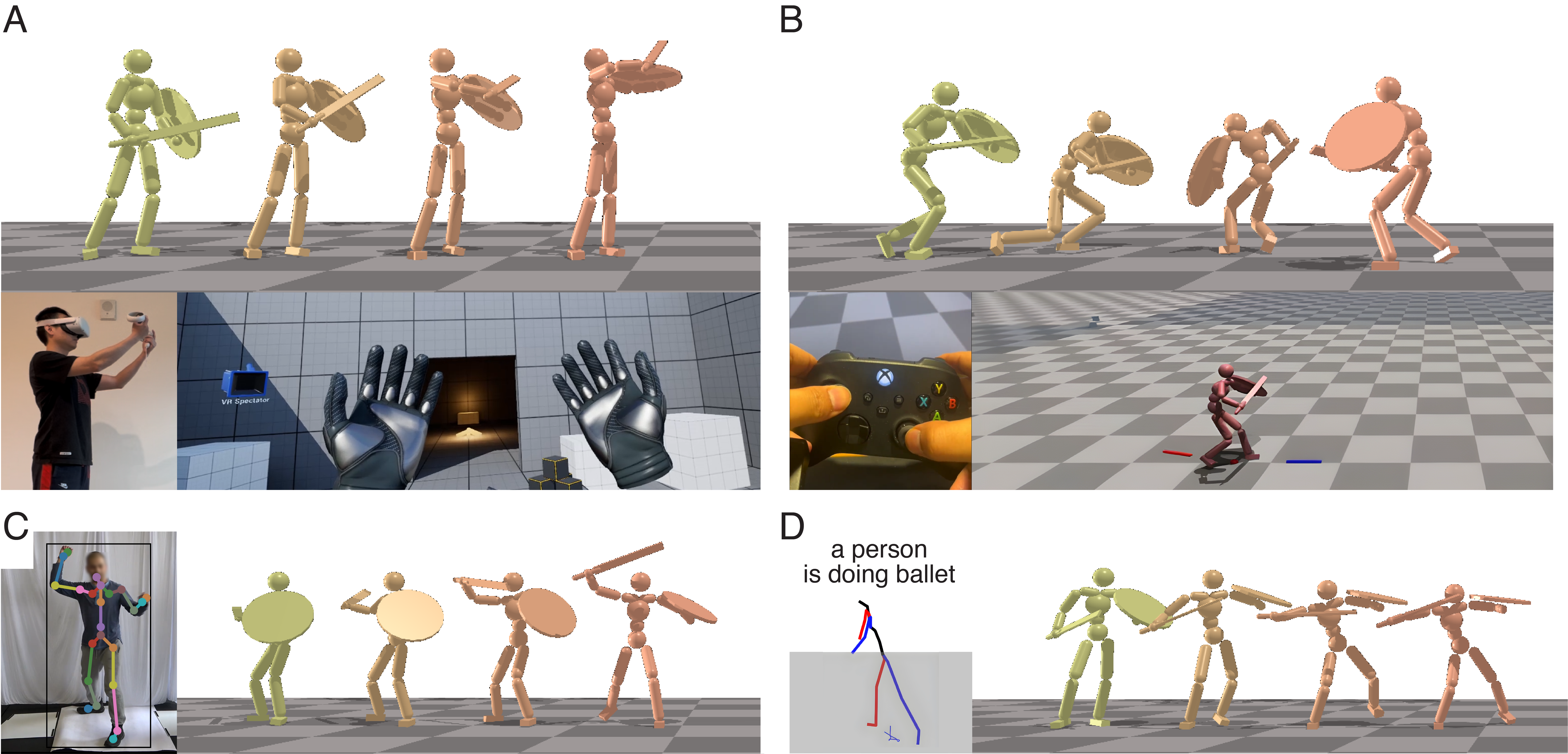}
  \caption{Showcases generated human motions from multi-modal inputs: (A) VR device, (B) joystick controller, (C) video, and (D) text. Our proposed method, Masked Humanoid Controller (MHC), can generate physically realistic motions from a wide variety of muli-modal directives. 
  }
  \label{fig:hero}
\end{figure}

\begin{figure}[tb]
  \centering
  \includegraphics[width=1.0\linewidth]{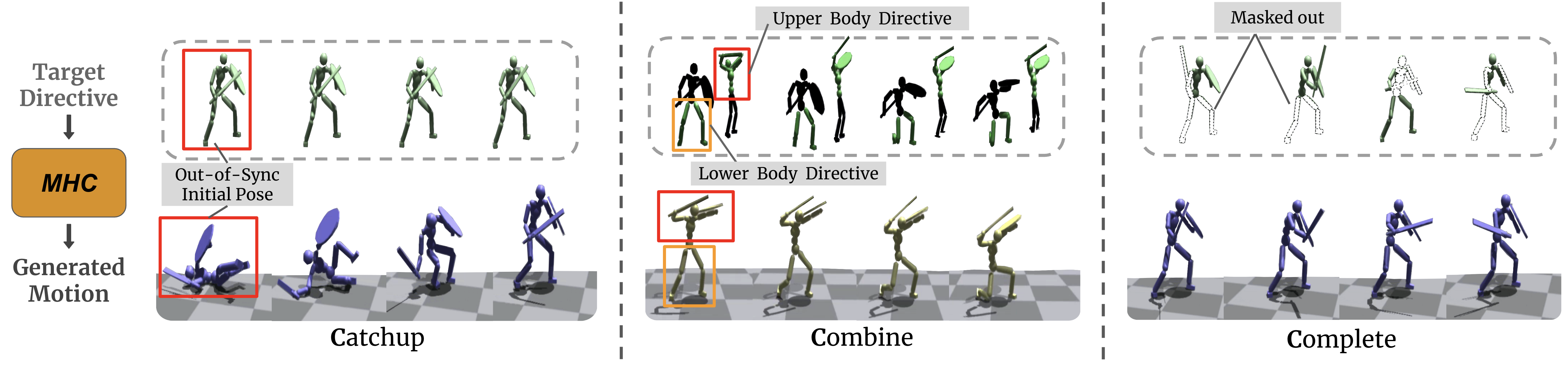}
  \caption{ Shows generated motions that illustrate the CCC capabilities. From left to right: MHC is able to generate motions that (1) adjust and \emph{catchup} starting from an out-of-sync pose, (2) imitate a target directive that \emph{combines} upper and lower body sub-segments from different motions, and (3) \emph{complete} the motion from under-specified directives as indicated by missing target outlines.
  }
  \label{fig:overview}
\end{figure}

A key aim of our work is to train a motion generator that can be directed through multiple input modalities such as video, text, and VR controllers as well as high-level goal specifications, without the need for fine-tuning. 
As an example, given limited demonstrations of \textit{running} and \textit{stationary waving}, we should be able to generate motions for a new behavior “running while waving” by directing the generator with high-level waypoints for navigation along with arm motion directives. Importantly, such waypoint directives are sparse in that they do not specify the joint-level details of the desired motion. Thus, the generator must be able to automatically fill-in-the-blanks to achieve natural looking motions that satisfy the sparse directives. In addition, the generator must be able to recover from failures, smoothly switch between different motions, and blend motions into novel combinations. We identify these crucial capabilities as: \textit{\textbf{Catch-up}}, \textit{\textbf{Combine}} and \textit{\textbf{Complete}} - (CCC) (Fig. \ref{fig:overview}).

The \textit{Catch-up} capability is required to support transitioning between different motion directives. This refers to resynchronizing to the directed motion from states that are currently inconsistent with that motion, e.g., failure states. This allows initiating motions from arbitrary fallen poses and seamlessly transitioning between different reference motions mid-way through execution.  The \textit{Combine} capability involves imitating novel combinations of upper and lower body movements by blending segments from different motion demonstrations; for instance, walking while waving. Finally, the \textit{Complete} capability allows for sparse, under-specified guidance signals; such as missing certain joint information, by tracking available cues and implicitly filling in missing details. Together, these capabilities enable multimodal directives across MoCap, text, video, joystick and VR. Additionally, they facilitate defining finite state machines to generate motions using high level user-specifications, as well as integration with data-driven planners to generate motions that achieve user-specified high-level goals.

To enable the CCC capabilities under a unified framework, we present the Masked Humanoid Controller (MHC). In summary, our main contributions are three-fold: (1) We identify three key missing capabilities in current humanoid controllers - catch-up, combine and complete - that are jointly-necessary for adaptable behaviors, (2) We propose the Masked Humanoid Controller (MHC) - a novel multi-objective imitation learning framework that utilizes augmented and masked demonstrations to enable these capabilities, (3) We integrate MHC with planning frameworks, enabling complex goal-driven behaviors using finite state machines (FSMs) and data-driven planners (DAC-MDPs \cite{Shrestha2020DeepAveragersOR}). The CCC capabilities are extensively validated by training an MHC on a diverse dataset of 87 motor skills\cite{Reallusion2022}. Additionally, we showcase MHC's zero-shot planning capabilities using a wide range of finite state machines as well as case studies on integration with DAC-MDPs for motion generation adhering to high-level goals.

\section{Related Work}
\label{sec:related}
\textbf{Learning from MoCap Data:} Leveraging MoCap data enables controllers to acquire complex behaviors with human-like motion quality. However, large and diverse MoCap datasets present challenges including scalable distillation and ensuring fluid transition between skills. Initial works train single-clip policies to mimic individual behaviors using tracking rewards \cite{peng_deepmimic_2018, Chentanez2018PhysicsbasedMC, ren_diffmimic_2023} or adversarial losses \cite{peng_amp_2021, Ho2016GenerativeAI, Lee2022HumaninspiredVI, Merel2017LearningHB, Xu2021AGA}. However, distilling these into a multi-clip controller remains computationally prohibitive \cite{wagener_mocapact_2023}. A more scalable alternative is to directly learn a multi-clip policy learning via reinforcement learning with tracking objectives \cite{Won2022PhysicsbasedCC, Won2020ASA, luo_perpetual_2023, Merel2018NeuralPM, Dou2023CASELC, wagener_mocapact_2023, Peng2022ASE}. However, tracking rewards alone is not enough to ensure smooth transitions between skills and failure recovery. Recent works augment training with adversarial losses to encourage natural motions during transitions \cite{Peng2022ASE, Tessler2023CALMCA} or define an explicit fail state recovery policy \cite{luo_perpetual_2023}. 

\textbf{Combination of motions:} Recent works have explored imitating combined motions, but require training individual policy for each new behavior pair \cite{Bae2023PMPLT, Xu2023CompositeML, Lee2022LearningVC}. Additionally, they rely on full motion oversight, lacking adaptability to partial guidance. On the completion front, some kinematic models can synthesize motions despite missing information \cite{Wang2020SynthesizingL3, Chen2022ExecutingYC}. However, these controllers are not grounded in physics, restricting their application.

\textbf{Under-specified Control:} Intuitive modalities like language, video, and VR provide important yet often under-specified means to direct motor skills. Existing works map some of these modalities to embedding spaces \cite{Juravsky2022PADLLP, Yao2023MoConVQUP, Lee2022HumaninspiredVI, Sun2023PromptPP} or key joint poses \cite{luo_perpetual_2023, Luo2023UniversalHM}. However, they handle a predefined sparisty types; adapting to new sparsity specifications like VR require expensive retraining \cite{Cern2023ANM, Du2023AvatarsGL}. While ambiguity is inherent in language and image-conditioned policies\cite{Huang2022PhysicallyPA, Yao2023MoConVQUP}, fine-grained control remains difficult as they do not allow low-level granularities like joint-level guidance. Overall a gap persists in controllers that can handle partial, sparse guidance with precision across modalities. Closing this gap can enable more intuitive control of reusable motor behaviors.

\textbf{Downstream Planning:} 
The acquired low-level control skills can support downstream tasks via a wide range of approaches including supervised fine-tuning for specialized behaviors \cite{Yao2023MoConVQUP},  reinforcement learning for new objectives \cite{merel_catch_2020, Luo2023UniversalHM, Peng2022ASE, Dou2023CASELC},  model predictive control for short-term horizons \cite{jiang2023hgap}, and finite state machines that encode behavioral logic \cite{Tessler2023CALMCA}. However, supervised fine-tuning remains restricted in flexibility to new tasks while reinforcement learning lacks sample efficiency whereas model predictive formulations are limited to short planning horizons. Additionally, the range of possible finite state machines largely depend on the flexibility of the underlying low-level controller. One solution is the use of data-driven planners like DAC-MDPs \cite{Shrestha2020DeepAveragersOR} which compile static experiences into approximated MDPs for fast optimization. While these methods enable zero-shot generalization, their integration with learned reusable motor skills remains relatively unexplored. Overall, leveraging low-level controllers to swiftly accomplish high-level goals remains an open challenge.

\section{Problem Formulation}
\label{sec:formulation}

Given a dataset of human motion demonstrations, our objective is to develop a motion generator that can produce directable, realistic combinations of those and similar motions in simulated physics environments. By directable we mean that the generator can be provided with directives that specify properties of the desired future motion. For example, the most detailed directive is to specify the exact future positions and velocities of each humanoid joint. Alternatively an under-specified directive  might provide just the desired hand positions in the future time window or just the torso velocity. For such under-specified directives, the motion generator should produce motion that is consistent with the provided directive, while also constraining to the space of natural motions. By allowing for under-specified directives we can 
support many input modalities for describing motion. For example, MoCap data provides fully-specified directives whereas joystick input may only specify the root velocity and arm motion.

\begin{figure}[t]
  \centering
  \includegraphics[width=\textwidth]{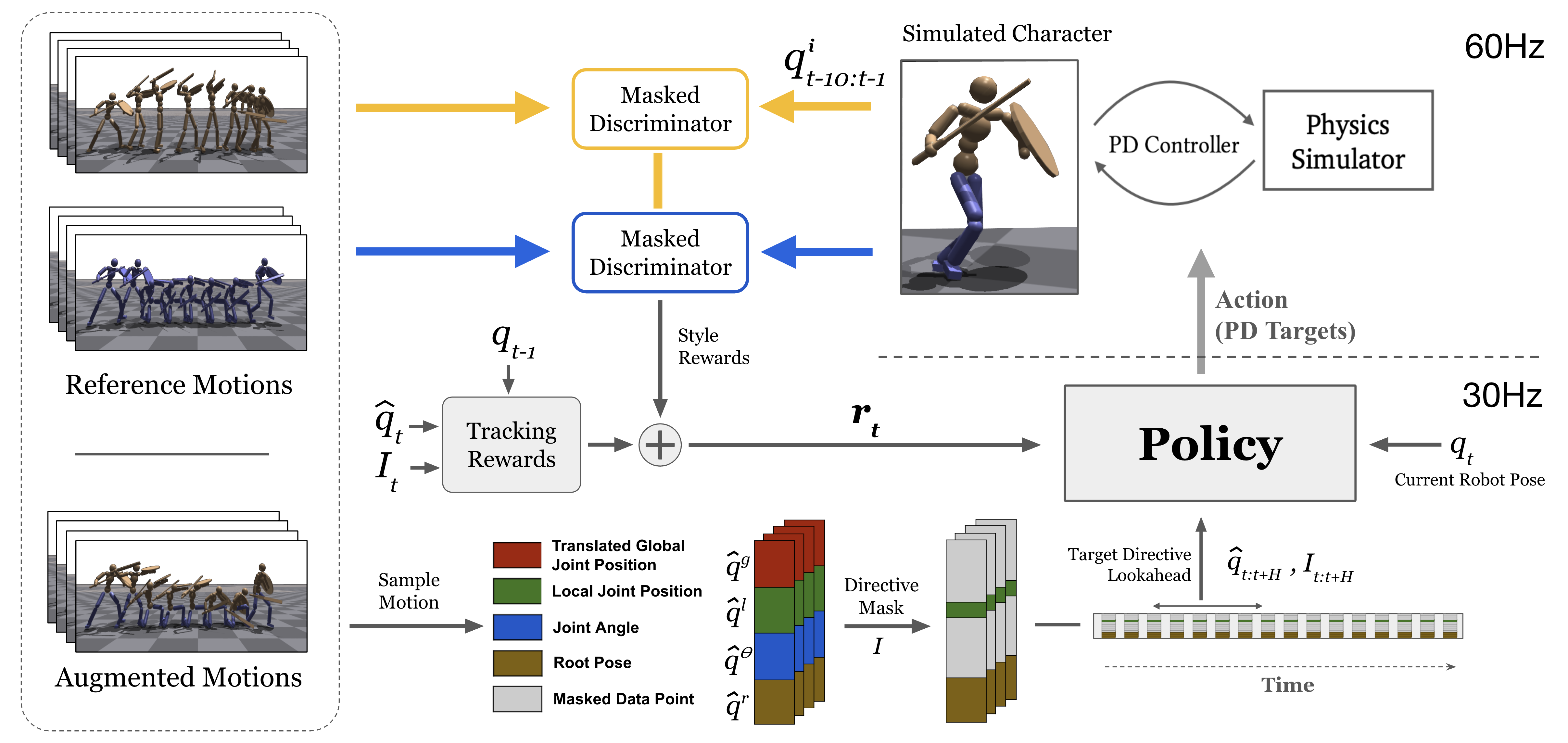}
  \caption{Illustrates the architecture and training details of the MHC framework, which consists of a controller and an ensemble of discriminators. Here the controller is trained to follow an augmented set of masked directives derived from the provided MoCap dataset. The controller gets feedback via tracking objective and style rewards generated by the ensemble of discriminators. Together they enable a directable policy to generate physically realistic motions capable of catching up, combining primitives, and completing motions from under-specified directives. (Figure layout adapted from \cite{Xu2023CompositeML})}
  \label{fig:block_diagram}
\end{figure}

More formally, a \emph{motion} is a sequence $q_{1:H}$ of multi-channel poses for a humanoid with $J$ joints. Each pose is a tuple $q_i = (q^{r}, q^{\theta}, q^{l}, q^{g})$, where $q^{r}$ specifies the position, orientation, linear and angular velocity of the root joint, $q^{\theta} \in \mathbb{R}^{J \times 6}$ \footnote{We use the 6 DoF rotation representation for orientation \cite{Peng2022ASE}.} is the 3D joint rotations, $q^{l} \in \mathbb{R}^{J \times 3}$ is the local joint positions relative to the root pose, and $q^{g} \in \mathbb{R}^{J \times 3}$ is the global joint position in the world coordinates. 

Motion directives are used to specify constraints on a desired motion to be generated. Specifically, in this work, a \emph{motion directive} $d$ is defined as a masked motion sequence represented by $d = (\hat{q}_{1:H}, I_{1:H})$, where $\hat{q}_{1:H}$ is a motion and $I_{1:H}$ is a sequence of binary masks such that $I_i$ indicates which dimensions of pose $q_i$ are selected as motion constraints. For example, a directive that only specifies future hand positions would select only the dimensions corresponding to hand positions in $\hat{q}^{l}$.

In this work, we focus on generating motions that are dynamically consistent with a physics-simulation environment. In particular, it is not sufficient to simply generate kinematic pose sequences that ``look right". For this purpose we consider motion generators that dynamically control the actuators of physically simulated humanoids. Specifically, as shown in Fig. \ref{fig:block_diagram} our motion generator takes as input the pose of the humanoid $q_t$ at time $t$ and the current motion directive $(\hat{q}_{t+1:t+H}, I_{t+1:t+H})$, which specifies the desired motion for the next $H$ steps. The output of the generator $a_t$ gives the set points for each joint of the simulated humanoid. These set points are then provided to a PD controller with fixed gains, which produces actuator torques for the physics simulator. Since, our motion generator controls a physical system and is conditioned on masked motion directives, we call it the Masked Humanoid Controller (MHC). 

In order to learn an MHC we assume access to a dataset of reference motions $\mathcal{M}$, which encompasses the types of motions that are needed for the application goals. Our goal is to learn an MHC capable of producing motions that are combinations and in-plane rotations of motions in $\mathcal{M}$. Specifically, we consider combinations of upper and lower body motions in $\mathcal{M}$ and denote the augmented dataset of all such combinations as $\mathcal{M}^{+}$. Note, however, that our learning framework can be instantiated for any meaningful type of combination. In addition, the MHC should be able to smoothly switch between motion sub-segments in $\mathcal{M}^{+}$, even when switching to an out-of-sync motion, which requires ``catching up". Note that ideally, the learned generator will generalize beyond motions in $\mathcal{M}^{+}$, but the degree of generalization is primarily an empirical question.

\begin{figure}[t]
  \centering
  \includegraphics[width=\textwidth,clip]{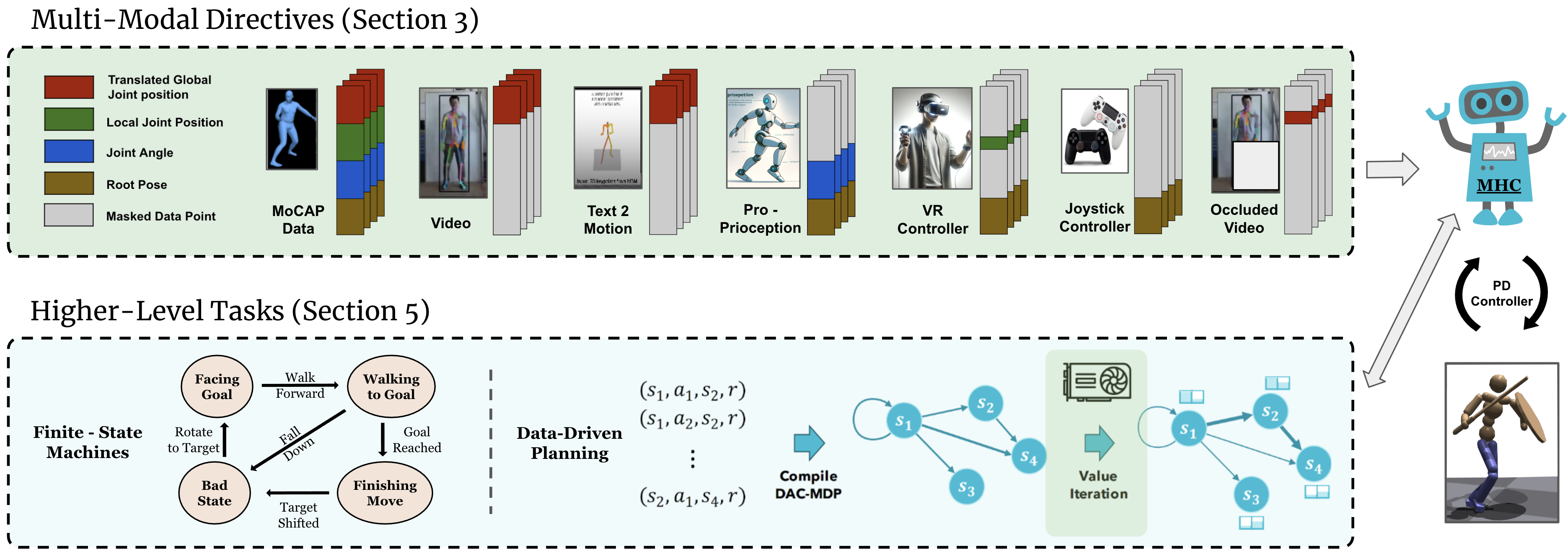}
  \caption{ Highlights the potential applications of MHC. [Top] The selective masking of the target directive allows MHC to represent various modalities of motion data under a single framework. These multi-modal inputs include MoCap, full or occluded video, joystick, VR controller among others. [Bottom] Similarly selective masking of target directive also allows us to treat the guiding signal itself as abstract actions. This enables straightforward integration with Finite State Machines and Data Driven Planning to allow zero-shot motion generation for higher-level task specifications.
  }
  \label{fig:downstream}
\end{figure}

Finally, the learned MHC should be able to produce motions based on different multi-modal sources of motion data. Each type of source will correspond to a particular type of directive whose mask identifies the motion constraints provided by that source. Figure. \ref{fig:downstream} illustrates the types of inputs and corresponding directive masks we consider in this work. For example, given human motion video, tracking software can be used to generate a directive $(\hat{q}_{1:H},I_{1:H})$ such that $\hat{q}$ only has components for $\hat{q}^g$ and $I$ selects only those components. Additionally, when joints are occluded, $I$ only selects the un-occluded joints.

\section{Masked Humanoid Controller}
\label{sec:mhc}
Because the MHC involves solving a complex physical control problem without supervision for the low-level actions, we formulate MHC training as reinforcement learning (RL). Training proceeds through episodes, where each episode initializes the humanoid in a physical environment with a specific directive. The reward during an episode indicates how well the  motion matches the target directive and reflects natural motion.

\subsection{Training Episodes}
Our goal is to create a distribution over episodes that will require the MHC to switch between motions in $\mathcal{M}^{+}$ based on directives corresponding to multimodal inputs. If an MHC achieves high reward on such a distribution, then it will have the desired CCC properties. Each episode is defined by a randomly sampled pair of initial humanoid pose and target directive. 

\textbf{Initial Pose Distribution:} The initial pose distribution is a mixture of a uniform distribution over poses in $\mathcal{M}^{+}$ and a distribution of fallen humanoid poses $p_{fall}(q)$, where the weight of $p_{fall}$ is 0.1 in our experiments. This choice forces the MHC to learn catchup to out-of-sync target directives, including recovery from fallen humanoid states. In addition, we apply a random in-plane rotation when generating a pose so that the MHC is robust to orientation changes.

\textbf{Target Directive Distribution:}  
For each episode we generate a target directive $(\hat{q}_{1:L}, I_{1:L})$, where $L$ is the length of the episode.  The motion $\hat{q}_{1:L}$ for the directive is generated by concatenating random length subsequences drawn from $\mathcal{M}^+$. In our experiments, $L=300$, corresponding to 10 seconds, and each sub-sequence length is uniform in the range from 120 to 240. This type of concatenation generally results in sharp and inconsistent motion transitions, forcing the MHC to learn catchup behavior. In addition, we also apply a random in-plane rotation to each sub-sequence so that the MHC is robust to orientation changes.  

The directive mask $I_{1:L}$ for an episode is generated by generating a mask $I_1$ for the first time step and using that mask for all other time steps (i.e. $I_t = I_1$ for all $t$).  We employ two types of masking: channel-level and joint-level masking (Fig. \ref{fig:downstream}). For, channel level masking, one or more of the available channels \((\hat{q}^{r}\), \(\hat{q}^{\theta}\), \(\hat{q}^{l}\), \(\hat{q}^{g})\) are selectively masked. The different channel combinations that we randomly choose between each episode are detailed in Fig. \ref{fig:downstream}. For example, all channels except $\hat{q}^{g}$ can be masked to represent the joint keypoint modality that can be generated from video via computer vision. Channel level masking allows us to represent modalities like video tracking, text2motion models, pro-prioception, and joystick controllers. 

Joint-level masking is a secondary level of masking that can be further applied to the joint position channels $(q^{l}, q^{g})$. Here we sample a percentage in $[0,100]$ and randomly select that percentage of joints to mask out from the target directive. This allows us to emulate partially occluded video motions as well as VR controllers which only have a limited number of keypoint sensors. In our experiments we sample masks by first randomly selecting a channel level mask followed by a 50\% chance of also applying joint-level masking.

\subsection{Reward Design} 

The RL training objective is to maximize the expected discounted episodic reward denoted by $\mathbb{E} \left[ \sum_{t=1}^{L} \gamma^{t-1} r_t \right]$ where $\gamma \in (0,1]$ is the discount factor and $r_t$ is the reward at timestep $t$. We define the reward at each step as the sum of a tracking reward $r^{tr}_t$, which encourages agreement with directives, a style reward $r^{\text{st}}_t$, which encourages natural looking motions, and an additional energy cost $c_t$, which encourages smooth motions: $r_t = 0.5r^{tr}_t + 0.5r^\text{st}_t - c_t$.


\textbf{Tracking Reward:} The tracking reward is defined to prefer generated motions based on how well they agree with the episode's directive at each time step. We find that learning is accelerated by using a reward function that prioritizes learning coarse motion characteristics before focusing on finer motion details. In particular, we define four reward terms in order of priority $r^{\text{h}}_t, r^{\text{o}}_t,r^{\text{v}}_t,r^{\text{l}}_t$ corresponding to matching the directive for root height, root orientation, root velocity and joint Euler coordinates, respectively. At each time step, a term is activated only if the higher priority rewards are above a threshold of 0.9. Specifically, the tracking reward function is given by: $r^{tr}_t = r^{\text{h}}_t + r^{\text{o}}_t +  r^{\text{v}}_t + r^{\text{l}}_t$
\begin{small}
\begin{align}
    r^{h}_t &=  e^{- m_h \cdot 8 ||q^h_t - \hat{q}^h_t||_{2}}\\
    r^{o}_t &= I(r^{h}_t > 0.9) \cdot e^{- m_o \cdot ||d(q^{o}_t - \hat{q}^{o}_t)||_{2}}\\
    r^{v}_t &= I(r^o_t > 0.9) \cdot e^{- m_v  \cdot ||q^{v}_t - \hat{q}^{v}_t ||_{2}}\\
    r^{l}_t &= I(r^{v}_t > 0.9) \cdot 
    \frac{1}{\underset{j \in J}{\sum} m_j}
    \underset{j \in J}{\sum} e^{- m_{j} \cdot 40 ||q^{j}_t - \hat{q}^{j}_t||_{2}}
\end{align}
\end{small}
where $J$ is the set of all joints, and $m_h$, $m_o$, $m_v$, $m_j$ are equal to 0 if the corresponding reward component depends on the pose information 
not selected by the episode's directive mask, and otherwise equal to 1. Thus, reward terms not relevant to the directive do not affect preferences over the generated motion.


\textbf{Style Reward:} We consider a multi-part style reward, inspired by \cite{Bae2023PMPLT}, with a distinct discriminator for different body parts. Specifically, we create 5 sets of joints from the whole body: \( \mathcal{J}_1 \mathcal{J}_2, \mathcal{J}_3, \mathcal{J}_4, \text{and } \mathcal{J}_5 \) corresponding to upper right, upper left, root, lower body, and full body, respectively. During RL training a discriminator $D_{\phi_k}$ is trained for each part set $\mathcal{J}_k$, conditioning only on joint information in $\mathcal{J}_k$. Training is done continuously throughout RL by sampling a pair of motions, one positive motion from $\mathcal{M}$ and one negative motion generated by the current MHC, and updating the parameters of each discriminator using the regularized loss function from \cite{peng_amp_2021, Bae2023PMPLT}. In our implementation, the input to the discriminators is a motion sub-sequence of length 10 so that the discriminator takes temporal dependencies into account. The corresponding reward component at time step $t$ for discriminator $k$ is $-\log(1 - D_{\phi_k}(q^k_{t-10:t})$. The overall style reward \( r^{\text{st}} \), is the average of these components across discriminators.

\textbf{Energy Cost:} To penalize large changes in the action across timesteps and large torques, the energy cost at time $t$ is given by
\begin{equation}
c_t = \sum_{j \in J} 0.01\cdot ||a^j_t-a^j_{t-1}||_1  + 0.0002 \cdot  ||\tau^j_t||_1 
\end{equation}
where $a^j_t$ is the action (set point) at time $t$ for joint $j$ and $\tau^j_t$ is the torque applied to joint $j$. The inclusion of this penalty is critical for avoiding high-frequency jitter, especially of the foot. 

\begin{figure}[tb]
  \centering
  \includegraphics[width=1.0\linewidth]{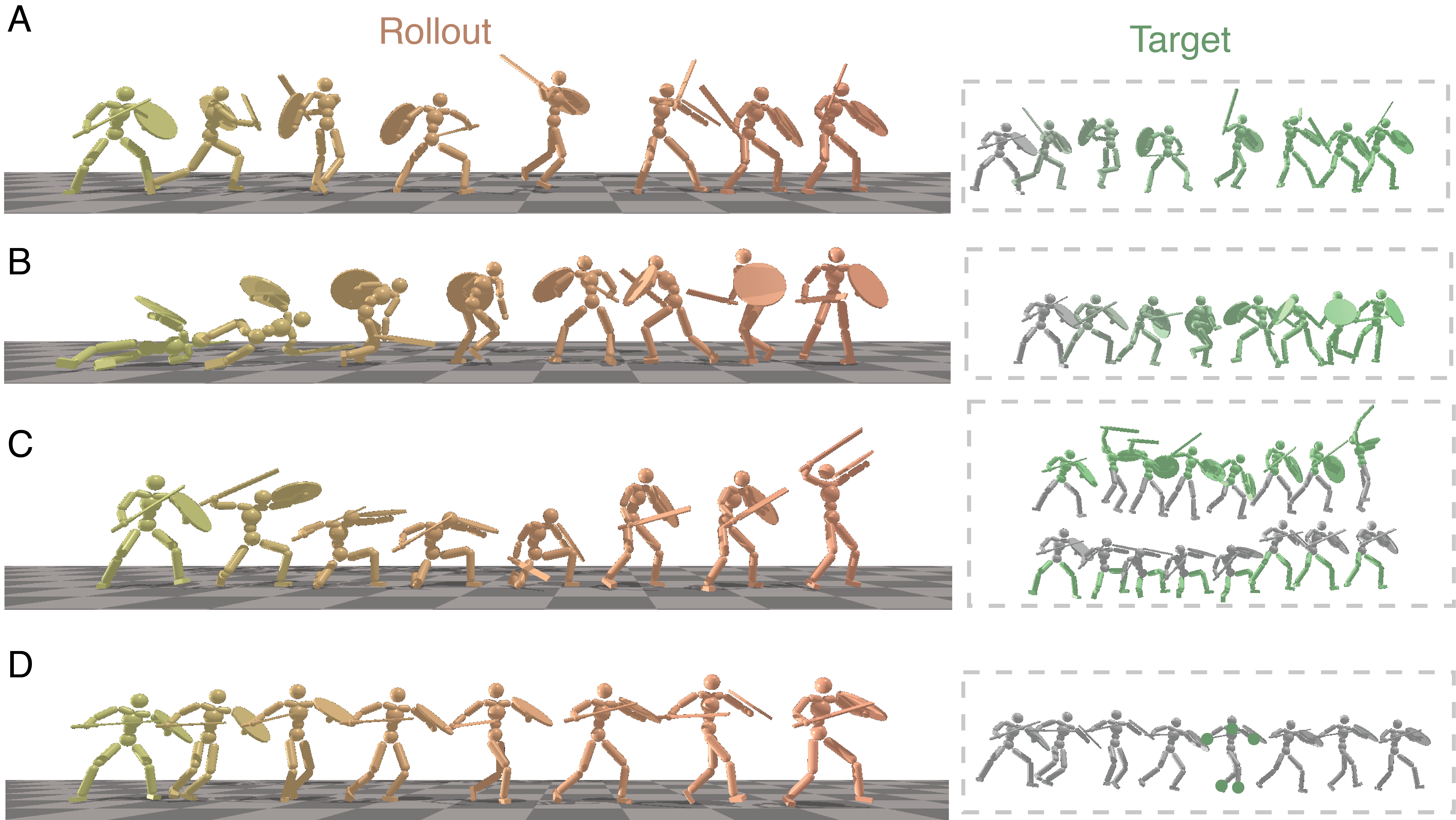}
  \caption{Illustrates generated motions corresponding to key CCC capabilities of MHC. The simulation (left) displays key-frames of humanoid following different motion directives (right). From top to bottom the simulated humanoid (A) follows an imitation target, (B) transitions from falling-down position to catch-up to the target directive, (C) imitates motion directive that combines upper-body and lower-body movements from distinct motions (D) completes the motion using only 3D joint positions of the head, hands and feets.
  }
  \label{fig:imitation}
\end{figure}

\section{Motion Generation for Higher-Level Tasks}

\label{sec:highlevel}

\begin{table}[t]
\caption{Results for the ASE baseline, the MHC, and MHC ablations for the imitation, catchup and combine experiments. The ablation differs across expeirments. For imitation we remove the style reward, for catchup we remove random pose initialization, and for combine we remove target motion augmentation (i.e. training on $\mathcal{M}$ rather than $\mathcal{M}^{+}$. The MHC outperforms the ASE baseline for all tasks across the Reallusion dataset $\mathcal{M}_{Real}$ and ASE rollout dataset $\mathcal{M}_{ASE}$. The ablation experiment demonstrate that each component is crucial and jointly enable the CCC capabilities. }
\label{tab:ccc_results}
\centering
\tiny
\renewcommand{\arraystretch}{2} 
\begin{tabular}{c|c|c|c|c|c|c|c|c|c|c|c|c}
\hline
\multirow{3}{*}{Method} & \multicolumn{4}{c|}{Imitation} & \multicolumn{4}{c|}{Catchup} & \multicolumn{4}{c}{Combine} \\
\cline{2-13}
 & \multicolumn{2}{c|}{$\mathcal{M}_{Real}$} & \multicolumn{2}{c|}{$\mathcal{M}_{ASE}$} & \multicolumn{2}{c|}{$\mathcal{M}_{Real}$} & \multicolumn{2}{c|}{$\mathcal{M}_{ASE}$} & \multicolumn{2}{c|}{$\mathcal{M}_{Real}$} & \multicolumn{2}{c}{$\mathcal{M}_{ASE}$} \\
\cline{2-13}
 & \(mpjpe \downarrow \) & \(Suc \uparrow \) & \(mpjpe \downarrow\) & \(Suc \uparrow\) 
 & \(mpjpe \downarrow \) & \(Suc \uparrow \) & \(mpjpe \downarrow\) & \(Suc \uparrow\) & \(mpjpe \downarrow\) & \(Suc \uparrow\) & \(mpjpe \downarrow\) & \(Suc \uparrow\) \\
\hline
ASE \cite{Peng2022ASE} &
123.51 & 0.6 & 100.28 & 0.77 & 
125.96 & 0.55 & 102.5 & 0.7 &  
210.75 & 0.17 & 197.94 & 0.25 \\
MHC (Ours) &
\textbf{ 51.05} & \textbf{0.92} & \textbf{56.23} & \textbf{0.97} &
\textbf{59.24} & \textbf{0.89 }& \textbf{63.46} & \textbf{0.98} &
\textbf{95.09} &\textbf{ 0.48} & \textbf{60.95} & \textbf{0.78} \\
MHC (abl) &
552.25 & 0.0 & 555.7 & 0.0& 
66.23 & 0.3 & 87.1 & 0.17 &
103.73 & 0.44 & 69.68 & \textbf{0.78}\\
\hline
\end{tabular}
\end{table}

We propose to integrate the learned MHC with a data-driven planning framework that allow for automatically generating a potentially large FSM for a user-defined higher-level task. Instead of specifying the FSM directly, the designer specifies: 1) a reward function $r^*$ and optimization objective (e.g. discounted total reward) corresponding to the higher-level task, 2) the set of possible directives $A$ for the MHC that can be used as dynamically selected actions to achieve the task, and 3) a mask over pose variables $S$ defining a state abstraction used to form the FSM states. For example, the user may specify a reward based on the distance to a goal location, specify directives corresponding to root velocity and orientation commands, and a state abstraction corresponding to localized goal location. Alternatively if certain types of arm motions were required, directives and the state abstraction could include selected sub-motions for the arm joints. 

We follow the data-driven planning framework of DAC-MDPs \cite{Shrestha2020DeepAveragersOR} which uses data collected from a dynamic system to construct and solve a Markov Decision Process (MDP) as shown in \ref{fig:downstream}. In our application, we collect trajectory data by randomly initializing the MHC in the environment and executing random sequences of directives from $A$ each trajectory stores the state abstraction at each time step along with the selected action directive. The DAC-MDP solution is a control policy that specifies for each abstract state, which directive in $A$ to execute. We refer the reader to \cite{Shrestha2020DeepAveragersOR} for DAC-MDP details. Importantly, the CCC properties of the MHC allow it to robustly execute the high-level actions selected by the DAC-MDP. It is worth noting that this approach is zero-shot in the sense that no fine-tuning or RL training is required to address a new high-level task. Rather, this framework uses efficient optimal planning to produce solutions.


\section{Experiments}
\label{sec:exp}
\begin{figure}[tb]
  \centering
  \includegraphics[width=1.0\linewidth]{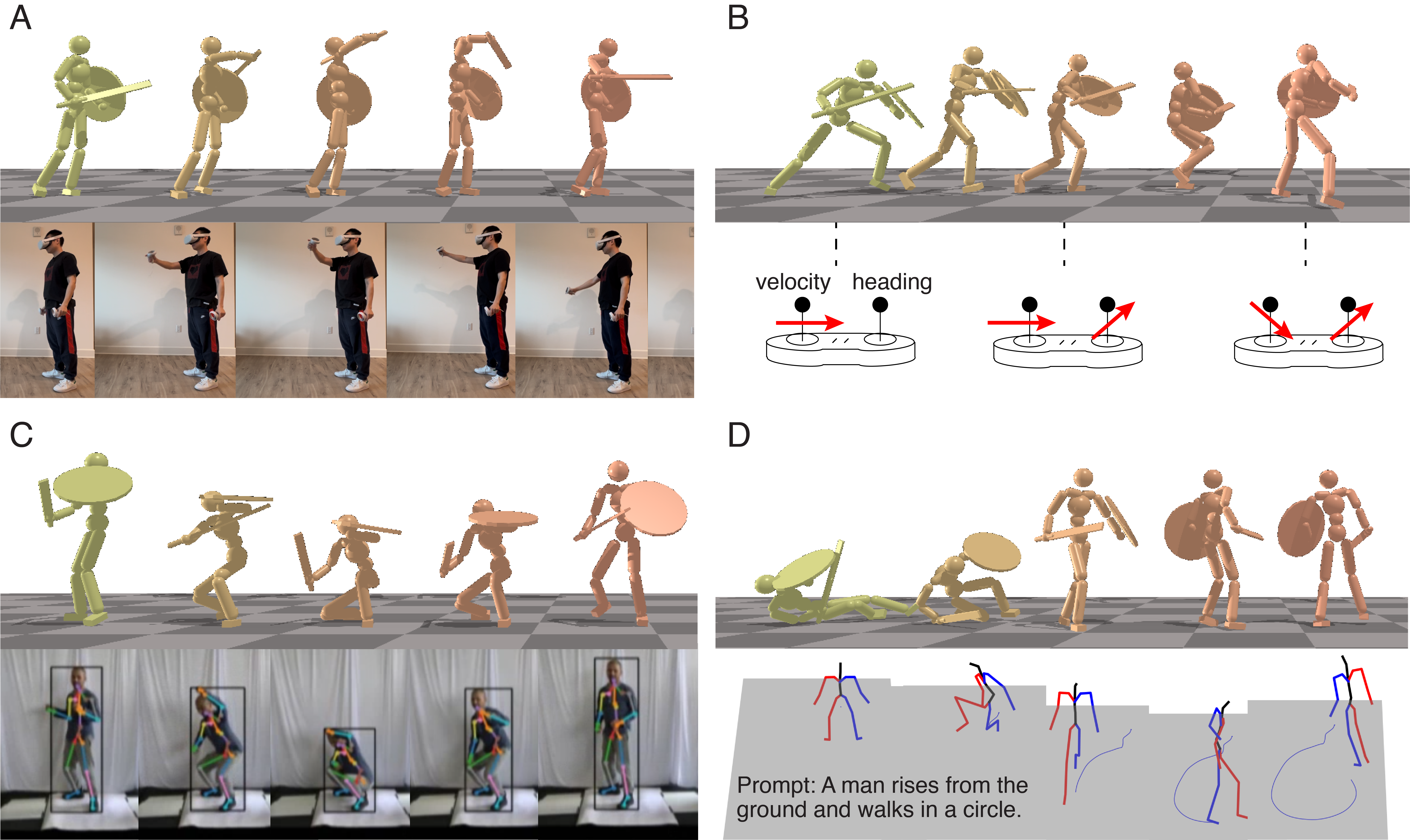}
  \caption{Illustrates qualitative results using keyframes for motion generation under multi-modal inputs such as (A) VR headset and controllers, (B) joystick controllers, (C) 3D joint positions derived from video and (D) text-to-motion generator. This highlights the versatility of MHC and its applications for motion generation directed by various modalities that may be noisy, under-specified.
  }
  \label{fig:modality}
\end{figure}

\begin{figure}[tb]
  \centering
  \includegraphics[width=1.0\linewidth]{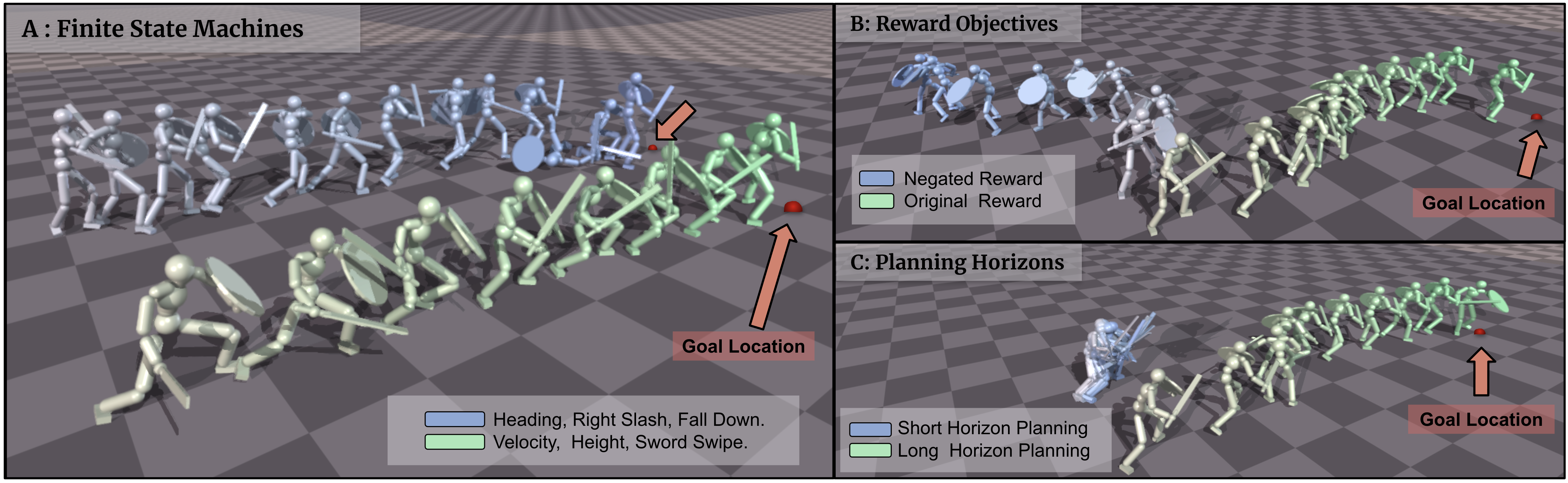}
  \caption{(A) Key frame visualization of FSMs for Go-To-Location task. FSM A1 (blue) generates a motion of walking towards the goal position while doing right sword slashes and ultimately falling down and recovering on reaching the goal. FSM A2 (green) generates a motion that swings its sword as it crouch-walks towards the goal while facing the other away. and swinging the sword along the way. (B) Visualization of FSM produced by DAC-MDP using original/negated reward functions for Go-To-Location task. FSM B1 (green) uses original reward function and reliably reaches the goal. FSM B2 (blue) uses negated reward function and avoids the goal. (C) Visualization of FSM produced by DAC-MDP for different discount factors. FSM C1 (blue) uses small discount factor and plans for short horizon. The generated motion swings the sword to get the immediate reward followed by episode termination. FSM C2 (green) use larger discount factor and plans for long horizon. The generated motion moves towards the goal to collect larger long-term reward. 
  }
  \label{fig:planning}
\end{figure}

We conduct experiments to empirically validate MHC's key capabilities to imitate a wide range of motion directives that necessitate the key capabilities of imitation, catchup, combination and completion (Fig. \ref{fig:imitation}). 
We then showcase motion generation from target directives derived from multiple modalities (Fig. \ref{fig:modality}) and case studies on motion generation for higher-level tasks. (Fig. \ref{fig:planning}). 

\textbf{Datasets and Baselines:} 
We train our MHC on a Reallusion MoCap dataset \cite{Reallusion2022, Peng2022ASE} of 87 motion clips ($\mathcal{M}_{Real}$). 
We also generate a test set $\mathcal{M}_{ASE}$ of 87 motions using a pre-trained ASE controller \cite{Peng2022ASE} by initializing it in a random pose and selecting a sequence of random ASE skills. We also use this pre-trained ASE controller as a state-of-the-art baseline for the Reallusion dataset. Specifically, ASE includes an encoder that maps target motions to skills, which allows us to evaluate ASE on fully specified directives. Note that, unlike MHC, the ASE controller does not support under-specified directives. Thus, in our comparisons, we always provide ASE with directives that specify all pose information.  

\begin{figure}[tb]
\centering
\includegraphics[width=\linewidth]{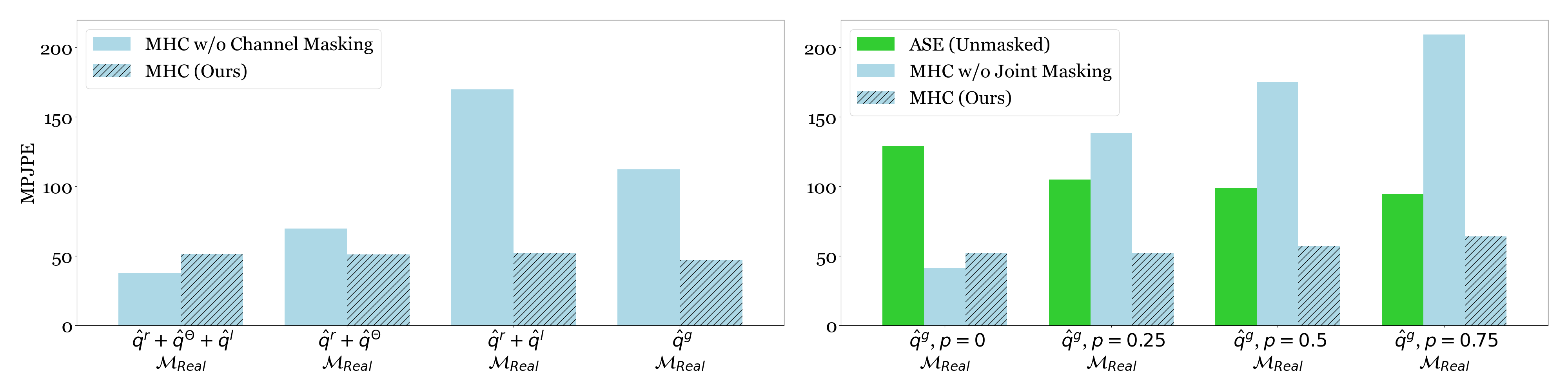}
\caption{(Left) Performance across different channel-level masks. We find that the MHC trained with directive masking retains its imitation performance across different variants of channel masks in contrast to the MHC trained without masking. (Right) Performance across different percentages of joint masking (0\% to 75\%). We see that the MHC shows stable performance as the amount of masking increases compared to the MHC trained without joint-level masking. We also see that the MHC significantly outperforms ASE, even though ASE is provided fully-specified (unmasked) directives. The performance of ASE varies across masking levels because we only evaluate the metric over unmasked joints. 
}
\label{fig:underspecification_results}
\end{figure}


\textbf{Metrics}
In our experiments we evaluate how well a generated motion matches the specified motion directive. As a metric of motion similarity we use the commonly used mean per-joint position error $E_{MPJPE}$ (in mm), which averages the root relative error of each humanoid joint.
Inspired by UHC \cite{Luo2022FromUH}, we also measure the success rate (\textit{Succ}) defined as following the reference motion with $< 10\%$ of failed frames, where a failed frame is one where maximum joint error is > 1m.  
For under-specified directives that mask joints, the success rate and  $E_{MPJPE}$ is measured only using the selected joints. 

\subsection{CCC Capabilities}
\textbf{Imitation:} We measure the imitation quality of the MHC and ASE baselines across both datasets $\mathcal{M}_{Real}$ and $\mathcal{M}_{ASE}$. Figure \ref{fig:imitation}A illustrates an example of the MHC imitating a target directive. 
To assess imitation performance, we use fully-specified target directives corresponding to the training and testing motions. The initial humanoid pose is set to match the initial pose of the target directive and then the controllers generate the motion using a sliding window lookahead of the target directive. As presented in Table \ref{tab:ccc_results}, MHC demonstrates superior performance compared to ASE on both datasets, as evidenced by the lower $E_{MPJPE}$ and higher success rate metrics. These results suggest that the MHC exhibits better imitation fidelity compared to the ASE baseline. 
We also consider an MHC ablation where we remove the sytle reward during training, using only tracking rewards and energy costs. Table \ref{tab:ccc_results} shows that the ablated MHC, denoted MHC(abl), fails to learn effective motion generation, even from these fully-specified directives. This finding underscores the critical role played by the style reward term in the successful training of MHC.



\textbf{Catchup:} To assess the catchup capabilities we initialize the MHC or ASE in a random pose drawn from the training or testing sets. We then run the motion generator with a directive that is derived by concatenating a pair of random sub-sequences from the training or testing sets. 
Figure \ref{fig:imitation}B showcases an example of the MHC successfully catching up to a target motion from a randomly initialized fallen pose, highlighting its adaptability and resilience. Table \ref{tab:ccc_results} presents a comparative analysis of MHC against ASE baseline and an ablation trained without random pose initialization. The evaluation is performed across both the training and test datasets, with the MHC consistently achieving superior $E_{MPJPE}$ and success rates compared to the baselines. Note that since catchup inherently involves non-synchronization with the directive, we loosen the success criteria so that a trial is considered successful if < 25\% of the frames fail.  These results provide strong evidence for MHC's enhanced capability in failure recovery and smooth catchup transitions, even when starting from perturbed states.

\textbf{Combine:} We now evaluate the ability to handle directives that combine upper and lower body joint subsets, as outlined in Section \ref{sec:mhc}. Here the target directive is generated by randomly sampling a pair of motion trajectories from either the training or testing sets and combining the upper and lower body movements from each motion.
Figure \ref{fig:imitation}C illustrates an example of the MHC successfully imitating a target directive derived by combining upper and lower-body movements from distinct motion sequences. 
Table \ref{tab:ccc_results} presents the quantitative results for imitating combined motions, where the motion pairs are sampled from $\mathcal{M}_{Real}$ and $\mathcal{M}_{ASE}$. We see that the MHC consistently outperforms both the ASE baseline and an ablated version of MHC. Here the ablation is trained without data augmentation i.e. $\mathcal{M}$ instead of $\mathcal{M}^{+}$. The ablation highlights the importance of data augmentation in enabling MHC to effectively work with directives that combine different body sub-segments from distinct motions.

\textbf{Complete:}
Finally, we evaluate MHC's ability to generate motions from under-specified target directives. We do this by repeating the setup for imitation task, this time with under-specified directives. As described in Section \ref{sec:formulation}, different choices of under-specification relate to different motion-data modalities. We start by evaluating channel level masks were we specify one of the following:  full pose information \((\hat{q}^{r} + \hat{q}^{\theta} + \hat{q}^{l})\), pro-prioception \((\hat{q}^{r} + \hat{q}^{\theta})\), pro-prioception with forward kinematics \((\hat{q}^{r} + \hat{q}^{l})\), and global joint keypoint positions \(\hat{q}^{g}\). 
Figure \ref{fig:underspecification_results} (left) shows that MHC can perform well for these channel level under-specification of directives. In a second experiment, we consider joint-level masking where we progressively mask 0\% to 75\% of the joints in \(\hat{q}^{g}\).
These directives correspond to potentially occluded keypoint trajectories from Video. Figure \ref{fig:underspecification_results} (right) shows that the MHC can retain its performance and generate motions with significantly lower $E_{MPJPE}$ as compared to the ASE baseline. This is notable since ASE is provided with all of the joint information, since it does not support under-specified directives. Finally, we condider an MHC ablation where we train without joint-level masking. We see that the ablated version performs poorly as the amount of masking increase. Figure \ref{fig:imitation}D illustrates an example of the MHC successfully imitating an under-specified target motion where only the 3D positions for the head, feet, and hands are provided. This highlights MHC's ability to generate realistic motion even from a diverse set of under-specified directives. 

\subsection{Multi-Modal Directives}

We qualitatively evaluate the MHC for different kinds of real-world input modalities for motions that the MHC was never trained on. We explore the following modalities: virtual reality devices, joystick controllers, 3D joint positions extracted from videos, and text-to-motion generation. In our experiments, we utilized the Meta Quest 2 VR system, comprising a headset and two controllers, to gather the orientation and 3D positions of both the headset and controllers (Fig. \ref{fig:modality}A). For joystick-based inputs, an Xbox controller was employed to manipulate the speed and direction of the velocity (via the left joystick and trigger), root orientation and height (via the right joystick and trigger), as illustrated in Fig. \ref{fig:modality}B. The estimation of 3D poses from video footage was achieved using MeTRAbs \cite{sarandi2021metrabs} (Fig. \ref{fig:modality}C), while generation of kinematic human motions from textual descriptions was achieved using T2M-GPT \cite{zhang2023generating} (Fig. \ref{fig:modality}D). We find that the learned MHC is versatile and can generate natural motions from under-specified directives derived from these input modalities. These results have important implications for real-world applications, where motion capture data may often be incomplete or noisy due to occlusions or sensor limitations. 


\subsection{Higher-level Task Specification}
We provide demonstrations of motion generation for higher-level tasks via integration of the MHC with FSMs and DAC-MDPs. 
For FSM integration we consider the task of navigating to a specified target goal location while executing different locomotion styles, such as running or crouch-walking, and performing various finishing moves, like taunting or falling down as shown in Figure \ref{fig:planning}A. We hand-coded simple FSMs to achieve these tasks, where each state of the FSM selects an appropriate directive for the MHC to execute. This includes under-specified directives corresponding to joystick commands and a fully-specified directives corresponding to the finishing moves. 

We integrate the MHC with DAC-MDPs by constructing a DAC-MDP using a dataset collected by generating motions using the MHC with directives derived from random joystick commands. The first high-level task (Figure \ref{fig:planning}B) is to simply reach a goal location. To do this, we define a DAC-MDP reward function that provides a reward at each step that is larger as the humanoid gets closer to the goal. The reward at the goal is a maximum of 1. The resulting FSM produced by solving this DAC-MDP results in motions that reliably reach the goal. To illustrate the zero-shot capabilities, we next give the DAC-MDP the negation of the first reward function, which should lead the agent to avoid the goal location. Figure \ref{fig:planning}B illustrates the resulting desired behavior. 

Finally, we illustrate the higher-level reasoning capabilities afforded via the DAC-MDP integration. As illustrated in Figure \ref{fig:planning}C, we adjust the  reward function and DAC-MDP termination condition so that the humanoid receives a positive reward for swinging the sword, which immediately terminates the episode. We then use the DAC-MDP to produce an FSM for a large discount factor 0.999 and a smaller discount factor 0.9. The large/small discount factors encourage planning over a long/short horizon. The best short horizon plan is to simply get the immediate reward of swinging the sword and then ending the episode, while the long-horizon plan is to instead go to the goal and collect the larger long-term reward. This is exactly the behavior produced by the two FSMs. 



\section{Summary} 

We highlighted three capabilities, Catchup, Combine, and Complete, that are jointly necessary for practical applications of  physically-realistic motion generation. Our proposed MHC is the first motion generator that achieves all three capabilities. This RL-based approach for training is highly versatile and can be applied to any available motion-capture dataset and multi-modal input directives. Importantly, the MHC is a dynamic motion generator in the sense that it actively generates motion in response to the current environment conditions. Finally, we demonstrated a straightforward integration of the MHC with data-driven planning to allow for zero-shot motion generation for higher-level tasks.  



\section*{Acknowledgements}
This work is supported by the NSF Award 2321851 and DARPA contract HR0011-24-9-0423.

\bibliographystyle{splncs04}
\bibliography{main}

\newpage

\appendix
\renewcommand{\thesection}{\Alph{section}}
\section{Appendix}

\subsection{Implementation Details} 
\label{sec:implementation}

The physics simulation is conducted using Isaac Gym \cite{Makoviychuk2021IsaacGH}, where the MHC runs at 30 Hz and the simulation runs at 60 Hz. The architecture of the MHC, which consists of a controller and an ensemble of discriminators, each implemented as a neural network. The \emph{controller} policy encodes the motion directive lookahead using a 3-layer perceptron with hidden dimensions of 1024 and 512. This encoding is then concatenated with the current pose of the humanoid and fed into another 3-layer perceptron with dimensions [1024, 1024] that serves as the policy head, outputting the action of the controller.
In practice, we implement the ensemble of \emph{discriminators} as a single discriminator with different wrappers. Each wrapper masks the observations to consider only a single set of joints. The discriminators are also 3-layer perceptrons with dimensions [1024, 512]. We use SiLU activations for all perceptrons.
For \emph{Training} the MHC is trained via the Proximal Policy Optimization (PPO) reinforcement learning algorithm \cite{Schulman2017ProximalPO} with fixed entropy. The training process takes 7 days on a single Nvidia A6000 GPU to obtain the final policy.

\newpage
\subsection{Extended results for High level task specification}

We further consider different variants of the go-to-location task and the heading task following \cite{Tessler2023CALMCA} to showcase the flexibility afforded by the MHC framework. The rewards for both tasks are defined following \cite{Tessler2023CALMCA}.

For the \textbf{heading task}, a random heading direction and a velocity direction are sampled as targets, and the FSM should be able to follow these directives. For example, "Head east while facing west." Furthermore, we can also specify the speed and height during these motions. "Run" maps to a speed of 2.5 m/s with a root height of 0.85m, while "crouch walk" refers to a root height of 0.4m with a speed of 1m/s. We evaluate the resulting motions using task-specific rewards, which reward matching the desired direction and heading. We find that the produced FSMs can reliably generate motions that match the higher-level heading task objectives.
 
For the \textbf{go-to-location task}, we consider different variations in movement time and finishing motion. Unlike \cite{Tessler2023CALMCA}, where only a particular skill can be requested as a finishing motion, our framework allows giving any kind of full or partial directive as a finishing motion. Here, we consider sword swing and taunt motions as finishing directives. The movement directives are chosen following the heading task. It is worth noting that our framework also allows for different upper body movements throughout the movement as well. Table \ref{tab:zero_shot_results} shows that FSMs using MHC generate motions that achieve high task rewards across all go-to-location task variants. 


\begin{table*}[h]
\caption{Quantitative evaluation of directional motion control (Heading) and zero-shot task solution. We consider two forms of locomotion Run and Crouch walk, each characterized by a different speed and style. We consider various finishing motions for the location task: 
\includegraphics[width=0.3cm]{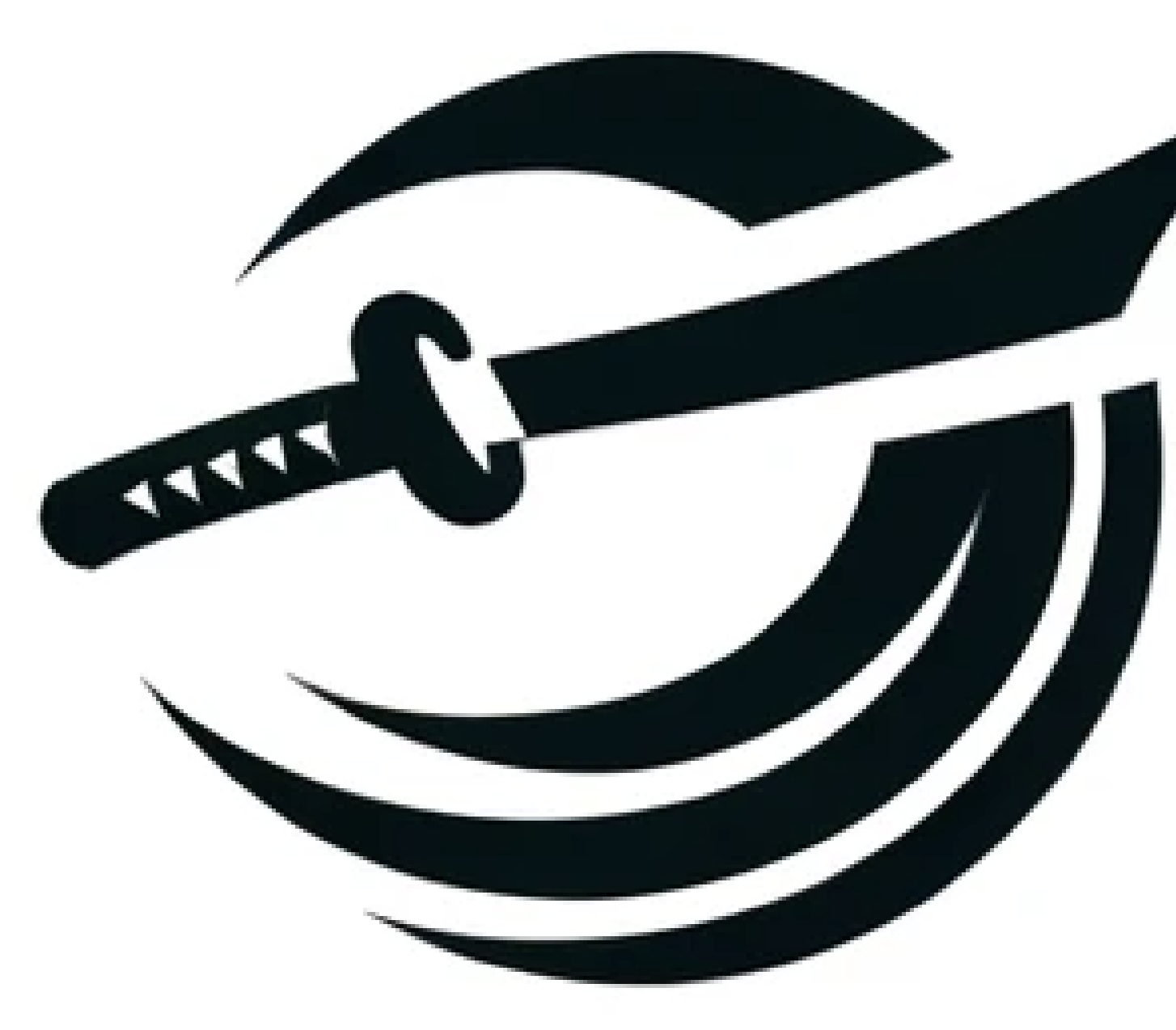} (Sword Swing) and 
\includegraphics[width=0.25cm,height=0.4cm]{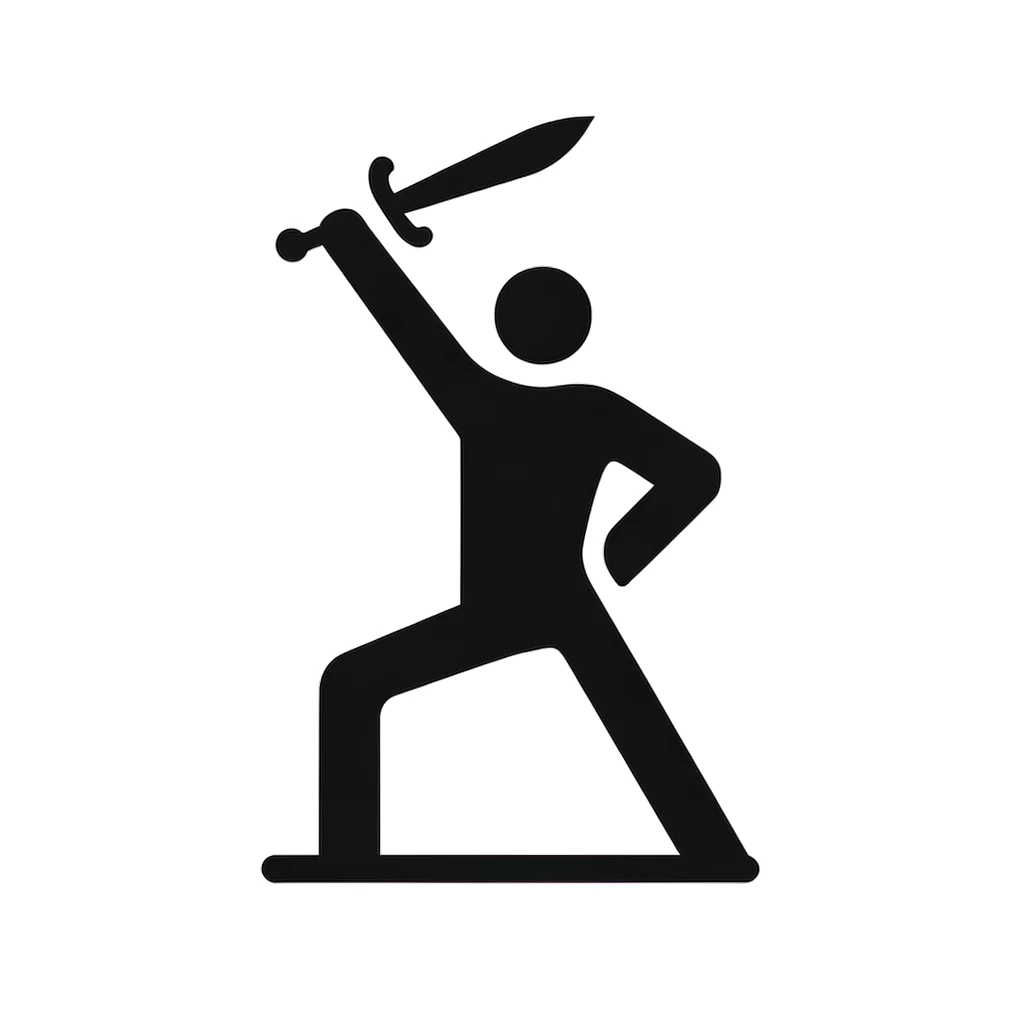} (Taunt).
}

\label{tab:zero_shot_results}
\centering
\scriptsize
\renewcommand{\arraystretch}{1.8} 
\begin{tabular}{l|c|c|c|c}
\hline
\textbf{Motion} & \multicolumn{2}{c|}{\textbf{Heading}} & \multicolumn{2}{c}{\textbf{Location}} \\
\cline{2-5}
 & \textbf{Style} & \textbf{Score} & \textbf{Ending} & \textbf{Score} \\
\hline
Run & 1 & 0.92 & \adjustbox{trim={.25\width} {.15\height} {.25\width} {0\height},clip,width=0.25cm, height=0.4cm}{\includegraphics{assets/icon_taunt.png}} & 0.98 \\

 &  &  & \adjustbox{trim={.05\width} {.05\height} {.05\width} {.05\height},clip,width=0.3cm}{\includegraphics{assets/sword_swing.png}}  & 0.98 \\
 
\hline
Crouch Walk & 0.94 & 0.91 & \adjustbox{trim={.25\width} {.15\height} {.25\width} {0\height},clip,width=0.25cm,height=0.4cm}{\includegraphics{assets/icon_taunt.png}} & 0.96 \\

 &  &  & \adjustbox{trim={.05\width} {.05\height} {.05\width} {.05\height},clip,width=0.3cm}{\includegraphics{assets/sword_swing.png}}  & 0.96 \\
\hline
\end{tabular}
\end{table*}

\end{document}